# Predicting Stress-strain Behaviors of Additively Manufactured Materials via Loss-based and Activation-based Physics-informed Machine Learning

Chenglong Duan[1] and Dazhong Wu[1,*]

[1] Department of Mechanical and Aerospace Engineering, College of Engineering and Computer Science, University of Central Florida, Orlando, FL 32816, USA

* Corresponding author
Email: dazhong.wu@ucf.edu

**Abstract**
Predicting the stress-strain behaviors of additively manufactured materials is crucial for part qualification in additive manufacturing (AM). Conventional physics-based constitutive models of materials have limitations such as oversimplification of material properties, narrow applicability to limited strain rates, while purely data-driven machine learning (ML) models often lack physical consistency and interpretability. To address these issues, we propose a physics-informed machine learning (PIML) framework that leverages physical laws in elastic and plastic regions to improve the predictive performance and physical consistency of ML models for predicting the stress-strain curves of additively manufactured polymers and metals. A polynomial regression model is first used to predict the yield point based on AM process parameters, then stress-strain curves are segmented into elastic and plastic regions. Two long short-term memory (LSTM) models are trained to predict two regions separately. For the elastic region, Hooke's law is embedded in the LSTM models for both polymers and metals; For the plastic region, Voce hardening law and Hollomon's law are embedded in the LSTM models for the polymer and metal, respectively. In addition, the PIML framework is implemented by two architectures, including the loss-based and activation-based PIML architectures, where the physical laws were embedded into the loss and activation functions, respectively. The performance of the two PIML architectures are compared with two LSTM-based ML models, three additional ML models, including a ridge polynomial regression model, an artificial neural network, and a transformer, and a physics-based constitutive model.  These models are built on experimental data collected from two additively manufactured polymers (i.e., Nylon and carbon fiber-acrylonitrile butadiene styrene) and two additively manufactured metals (i.e., AlSi10Mg and Ti6Al4V). Experimental results have demonstrated that two PIML architectures consistently outperform three ML baseline models and one physics-based constitutive model. The segmental predictive model with activation-based PIML architecture achieves the best overall performance, with the lowest MAPE of 10.46±0.81% and the highest $R^2$ of 0.82±0.05, indicating superior predictive performance and stable generalization across materials.

## 1. Introduction
Additive manufacturing (AM) is a transformative technology that can fabricate parts with complex geometries [1, 2], lightweight structures [3, 4], and functional components [5, 6] across aerospace [7, 8], biomedical [9, 10], and automotive industries [11, 12]. By fabricating parts layer-by-layer, AM achieves much higher design freedom than subtractive manufacturing, allowing for rapid prototyping [13-15]. However, AM has not been widely adopted in critical applications due to inconsistent mechanical performance of additively manufactured parts [16]. Specifically, the mechanical performance of additively manufactured parts, particularly their stress-strain behaviors, remains highly sensitive to material variability, part geometry, and process condition [17-22]. Therefore, part qualification techniques are used to verify the stress-strain behaviors of additively manufactured parts satisfy the required design specifications to ensure these parts can be consistently reproduced, scaled for mass production, and reliably used in critical applications [23, 24].

Conventional AM part qualification techniques rely heavily on destructive and non-destructive testing. These techniques are expensive, time-consuming, and difficult to scale. Therefore, the physics-based constitutive models (e.g., plasticity and viscoelastic models) [25-28] and data-driven machine learning (ML) models [29-34] have been used to predict the stress-strain behaviors of additively manufactured parts. However, physics-based constitutive models require extensive parameter calibration based on prior knowledge and often oversimplify material properties. In many cases, physics-based constitutive models are sensitive to initial parameter selection and exhibit limited generalizability across different materials. For the data-driven ML models, although they can capture the complex, nonlinear relationship between input and output automatically, they typically lack inherent physical constraints. As a result, the predicted stress-strain behaviors may not be consistent with the actual stress-strain behaviors.

Recent advances in physics-informed machine learning (PIML) offer a potential solution by embedding physical laws directly into the training process of a ML model. While the effectiveness of PIML has been demonstrated in fluid dynamics [35, 36], heat transfer [37, 38], and other fields [39, 40], To the best of our knowledge, no studies have been reported on predicting the stress-strain behaviors of additively manufactured parts with PIML architecture. One challenge associated with applying PIML to solve problems in this field is the complex deformation behavior of materials. Specifically, the elastic region is governed by a simple and linear elasticity law (e.g., Hooke's Law), but the plastic region is governed by a complex and non-linear hardening law (e.g., Voce hardening law). Moreover, the plastic deformation mechanisms of additively manufactured polymers and metals differ substantially due to their underlying microstructural and molecular structures. Therefore, a material-specific and region-specific PIML framework is highly desirable to predict the stress-strain behaviors of additively manufactured polymers and metals.

Furthermore, although physics can be embedded into ML models through various mechanisms, embedding physics directly into the loss function is the most common approach in existing literature. For example, most of the existing loss-based PIML architectures integrate physics into a loss function with a learnable physics weighting coefficient. Alternatively, the activation-based PIML architecture enforces physics as hard constraints within a ML architecture. By integrating physics into the activation function of the output layer, it ensures consistent predictions at every time step and reducing reliance on the physics weighting coefficient in loss-based PIML architecture.

To address the aforementioned issues, a material-specific and region-specific PIML framework was developed to predict the stress-strain behaviors of additively manufactured polymers and metals. Specifically, two polymer and two metal datasets were used to demonstrate the proposed framework, including Nylon (Ultimaker) and carbon fiber-acrylonitrile butadiene styrene (CF-ABS, Push Plastic) samples fabricated by FFF, AlSi10Mg (3D Systems, Inc) and Ti6Al4V (3D Systems, Inc) samples fabricated by laser powder bed fusion (L-PBF). By firstly segmenting a stress-strain curve into an elastic region and a plastic region based on the predicted yield point, the material-specific and region-specific physical laws are embedded into the PIML architecture. Specifically, a loss-based and an activation-based PIML architectures were developed by embedding the physical laws into a loss function or an activation function, respectively. We also compared the performance of the proposed PIML models with two LSTM-based ML models (i.e., non-segmental and segmental predictive model without PIML), three additional ML models (i.e., ridge polynomial model, artificial neural network (ANN), and transformer), and one physics-based constitutive model to further demonstrate the effectiveness of the PIML framework. The main contributions of this study are summarized as follows:

1. A PIML framework that integrates physical laws (i.e., constitutive models) into ML models was developed to predict the complex stress-strain behaviors of additively manufactured polymers and metals.
2. Two PIML architectures, including a loss-based and an activation-based PIML architectures, were developed by embedding physical laws into the loss and activation functions, respectively.

3. The loss-based and activation-based PIML architectures were demonstrated on two additively manufactured polymer datasets and two additively manufactured metal datasets through a 5-fold cross validation, outperforming two LSTM-based models, three additional ML models, and a physics-based constitutive model.

## 2. Material and methods
### 2.1 PIML framework

Fig. 1 illustrates the three-step PIML framework for predicting the stress-strain behaviors of additively manufactured polymers and metals. In step 1, the predicted yield strain ($\varepsilon_{py}$) and yield stress ($\sigma_{py}$) are first obtained from process parameters in each fold. Since the relationship between process parameters and yield strain or yield stress is fundamentally a scalar-to-scalar mapping, it can be effectively modeled using a simple regression model. To determine an optimal model, multiple regression and ML models, including extreme gradient boosting (XGBoost) model, ridge polynomial regression model, support vector regression (SVR) model, and ANN, were evaluated. Table 1 shows the average mean absolute percentage error (MAPE) between predicted and actual yield strain and yield stress for the Nylon, CF-ABS, AlSi10Mg, and Ti6Al4V samples using four different models. Among the four models, the ridge polynomial regression model achieves the best predictive performance, with an average MAPE of 5.51% for yield strain and 7.71% for yield stress, respectively. Therefore, a ridge polynomial regression model is used to predict the yield point.

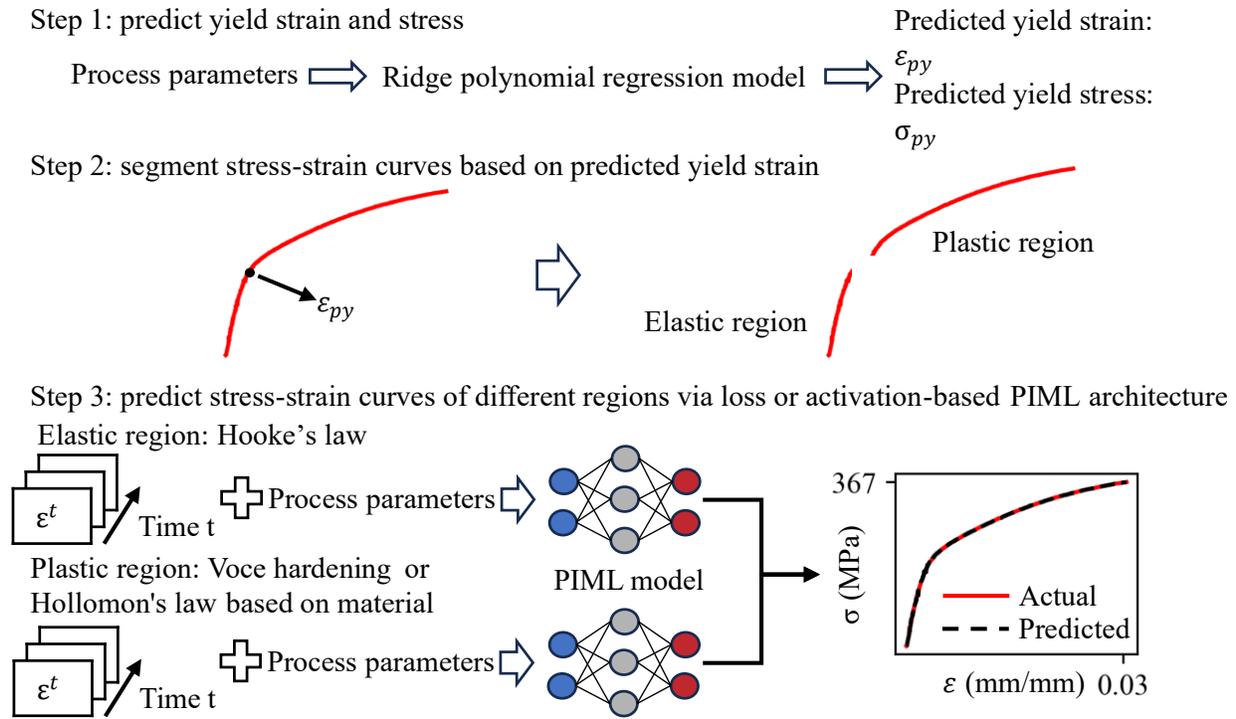

Fig. 1. Overview of the PIML framework.

Table 1. Average MAPEs between predicted and actual yield strain and yield stress of the Nylon, CF-ABS, AlSi10Mg, and Ti6Al4V samples.

|  | XGBoost | SVR | ANN | Ridge polynomial |
|---|---|---|---|---|
| Predicted yield strain ($\varepsilon_{py}$) | 7.79% | 8.12% | 315.63% | **5.51%** |
| Predicted yield strain ($\sigma_{py}$) | 9.15% | 9.04% | 13.56% | **7.71%** |

In step 2, the predicted yield strain is used to segment each stress-strain curve into an elastic region and a plastic region, allowing the subsequent PIML architectures to focus on the material-specific and region-specific physical laws. In step 3, the stress-strain curves in elastic and plastic regions are predicted using either the loss-based or activation-based PIML architectures. Each stress-strain curve is split into an input matrix and a corresponding output sequence. Each input matrix consists of a sequence of strain values ($\varepsilon^t$, time-variant) along with constant process parameters (time-invariant) concatenated at each time step. Each output sequence consists of the stress values at each time step ($\sigma^t$, time-variant). After obtaining two stress sequences for the elastic region and plastic region, these two sequences are combined to obtain the final predicted stress-strain curve. The detailed mechanism of loss-based and activation-based PIML models will be introduced in Section 2.2.

## 2.2 Embedding physical laws into loss function or activation function

To implement the PIML framework, we developed two PIML architectures, including loss-based and activation-based PIML architectures. Fig. 2 illustrates the loss-based PIML architecture for both elastic and plastic regions, the physical law is embedded into an additional physics loss function. Specifically, the total loss $L_{total}$ is composed of three terms: a data-driven loss $L_{data-driven}$, a physics loss $L_{physics}$, and an auxiliary loss $L_{auxiliary}$:

$$L_{total} = L_{data-driven} + \lambda_{physics} * L_{physics} + L_{auxiliary} \tag{1}$$

where $\lambda_{physics}$ is a learnable physics weighting coefficient of $L_{physics}$. At each time step $t$, a sequence of strains and corresponding process parameters are fed into the LSTM model. Then, based on (elastic or plastic) and material type (polymer or metal), the LSTM model outputs the predicted stress ($\sigma_p^t$) and the physics variables required by the physics loss $L_{physics}$. For the elastic region of both polymer and metal samples, the Young's modulus ($E_p^t$) is predicted, the Hooke's law [41] is integrated into $L_{physics}$:

$$L_{physics} = \frac{1}{N}\sum_{i=1}^{N}\left([\sigma_p^t]_i - [E_p^t * \varepsilon^t]_i\right)^2 \tag{2}$$

For the plastic region of the polymer samples, the stress amplitude $a_p^t$ and strain-hardening rate $b_p^t$ are predicted, the Voce hardening law [42] is integrated into $L_{physics}$:

$$L_{physics} = \frac{1}{N}\sum_{i=1}^{N}\left([\sigma_p^t]_i - \left[\sigma_{py} + a_p^t(1 - e^{-b_p^t(\varepsilon^t-\varepsilon_{py})})\right]_i\right)^2 \tag{3}$$

where $\sigma_{py}$ and $\varepsilon_{py}$ are predicted yield stress and strain in step 1 of the PIML framework.

For the plastic region of the metal samples, the strength coefficient $K_p^t$ and strain-hardening exponent $n_p^t$ are predicted, the Hollomon's law [43] is integrated into $L_{physics}$:

$$L_{physics} = \frac{1}{N}\sum_{i=1}^{N}\left([\sigma_p^t]_i - \left[\sigma_{py} + K_p^t(\varepsilon^t - \varepsilon_{py})^{n_p^t}\right]_i\right)^2 \tag{4}$$

For $L_{data-driven}$, it is the mean squared error (MSE) between the predicted stress $\sigma_p^t$ and the actual stress $\sigma^t$ to enforce their agreement:

$$L_{data-driven} = \frac{1}{N}\sum_{i=1}^{N}\left([\sigma_p^t]_i - [\sigma^t]_i\right)^2 \tag{5}$$

For $L_{auxiliary}$, this term acts as a constraint that keeps the learnable physics weight $\lambda_{physics}$ near 1, ensuring balanced contributions between the $L_{data-driven}$ and $L_{physics}$:

$$L_{auxiliary} = \alpha(\lambda_{physics} - 1)^2 \tag{6}$$

where $\alpha$ is equals to 0.1. This scaling coefficient $\alpha$ is set up to maintain the auxiliary regularization at a comparable numerical scale to the other loss terms, ensuring it only stabilizes the learnable physics weigh $\lambda_{physics}$ without influencing the overall optimization trend of $L_{total}$.

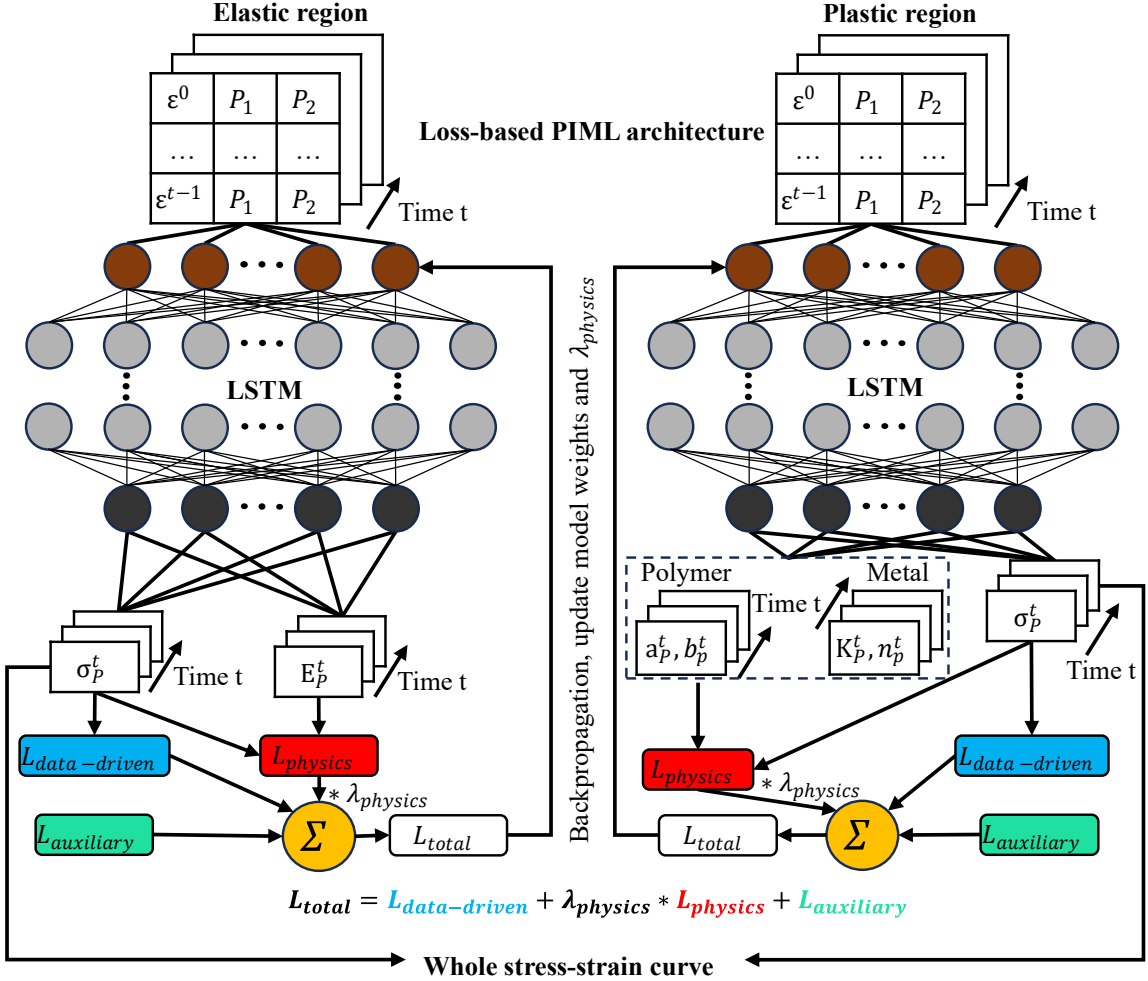

Fig. 2. Loss-based PIML architecture.

Fig. 3 illustrates the activation-based PIML architecture for both elastic and plastic regions. Instead of embedding physical law in the loss function, the activation-based PIML architecture embeds physical law directly into the activation function of the output layer. At each time step $t$, the LSTM model receives a sequence of strains and corresponding process parameters. Then, based on region (elastic or plastic) and material type (polymer or metal), the LSTM model first outputs the physics variables required by the activation function $f$, after which the activation function $f$ transforms these physics variables into the final predicted stress ($\sigma_p^t$). For the elastic region of both polymer and metal samples, Young's modulus ($E_p^t$) is predicted, the Hooke's law is integrated into activation function $f$ to transform $E_p^t$ into $\sigma_p^t$:

$$\sigma_p^t = f(E_p^t) = E_p^t * \varepsilon^t \tag{7}$$

For the plastic region of the polymer samples, the stress amplitude $a_p^t$ and strain-hardening rate $b_p^t$ are predicted, the Voce hardening law is integrated into activation function $f$ to transform $a_p^t$ and $b_p^t$ into $\sigma_p^t$:

$$\sigma_p^t = f(a_p^t, b_p^t) = \sigma_{py} + a_p^t(1 - e^{-b_p^t(\varepsilon^t - \varepsilon_{py})}) \tag{8}$$

For the plastic region of metal samples, the strength coefficient $K_p^t$ and strain-hardening exponent $n_p^t$ are predicted, Hollomon's law is integrated into activation function $f$ to transform $K_p^t$ and $n_p^t$ into $\sigma_p^t$:

$$\sigma_p^t = f(K_p^t, n_p^t) = \sigma_{py} + K_p^t(\varepsilon - \varepsilon_{py})^{n_p^t} \tag{9}$$

$L_{physics}$ is the MSE between the predicted stress $\sigma_p^t$ and the actual stress $\sigma^t$ to enforce their agreement:

$$L_{physics} = \frac{1}{N}\sum_{i=1}^{N}\left([\sigma_p^t]_i - [\sigma^t]_i\right)^2 \tag{10}$$

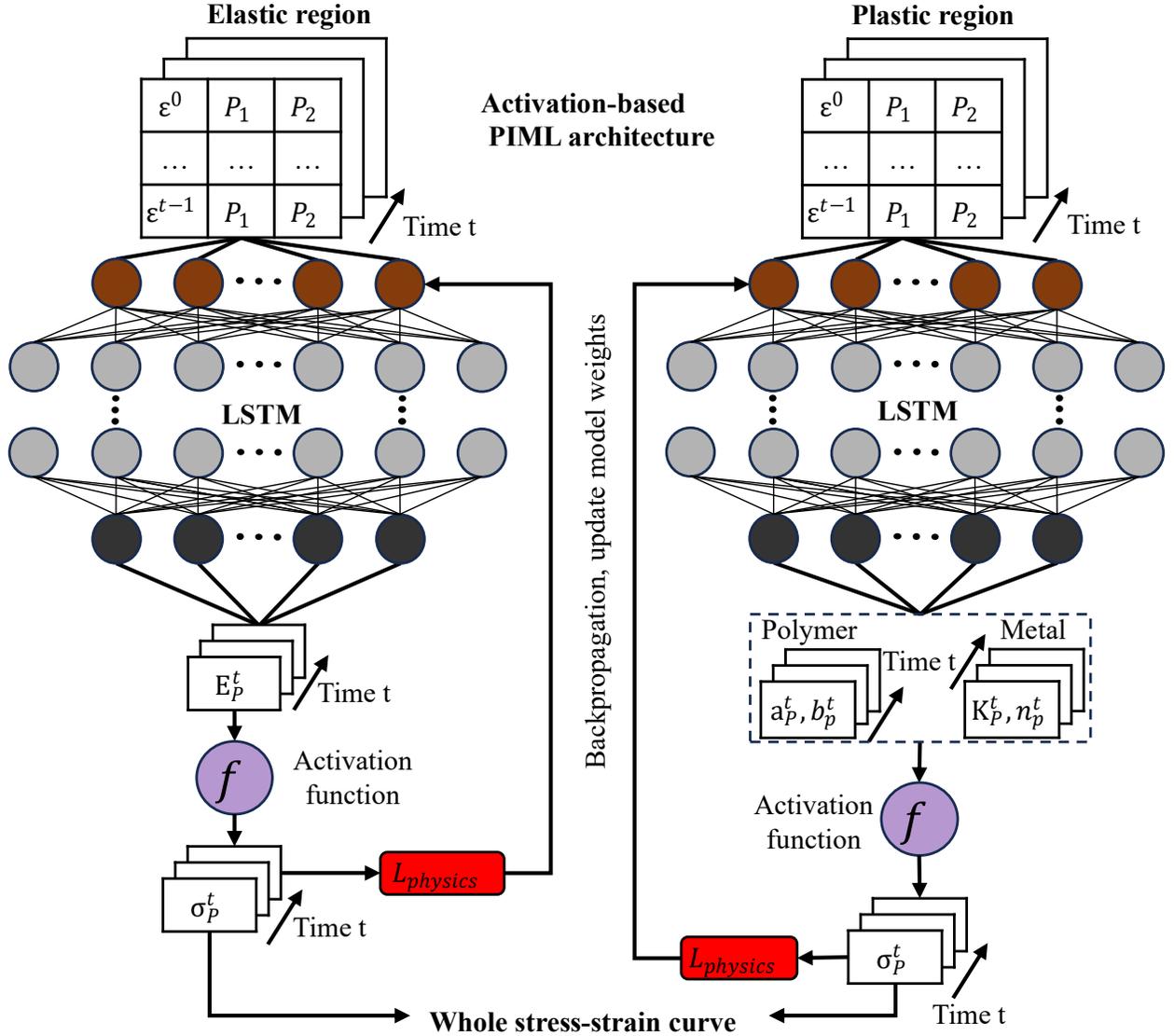

Fig. 3. Activation-based PIML architecture.

### 2.3 Machine learning and physics-based constitutive models
### 2.3.1 Machine learning models

For the LSTM model used in the PIML framework, it originally evolved from the recurrent neural network [44]. By incorporating a gate-controlled memory cell, LSTM is designed to handle the long-term dependency problem often encountered in sequential tasks [45]. The core component is the cell state, which acts as a memory capable of storing information over extended time steps. Three gates include forget gate, input gate, and output gate, controlling the flow of information into and out of the cell state. The forget gate determines how much of the previous cell state should be retained, while the input gate decides the degree to which new candidate information updates the current cell state. Finally, the output gate governs how much of the updated cell state is revealed as the hidden state at the current time step. More details about LSTM can be found in the literature.

In this study, the LSTM model was built using PyTorch and was trained on a computer equipped with an NVIDIA GeForce RTX 4090 GPU, an Intel i9 14900KF CPU, and 32 GB of RAM. The model consists of one LSTM layer that contains 32 hidden units, followed by a fully connected layer that maps the final hidden state to a single stress output. The learning rate is 0.01, the batch size is 1, the training epochs are 10 for all datasets. Moreover, since stress-strain curves are time-series data, the sequence length is another crucial hyperparameter when designing the LSTM input structure. For a sequence length $n$, the input from the $n$ sequential data points (strain and parameters) are combined to predict the $(n+1)$-th stress. Specifically, for the elastic region, the sequence length is 2, 2, 5, 10 for Nylon, CF-ABS, AlSi10Mg, Ti6Al4V dataset, respectively; for the plastic region, the sequence length is 10, 10, 5, 10 for Nylon, CF-ABS, AlSi10Mg, Ti6Al4V dataset, respectively.

To better demonstrate the effectiveness of the proposed PIML framework based on the LSTM model, three additional ML models, including a ridge polynomial regression model, an ANN, and a transformer, are included for comparison. The baseline ML models are introduced to provide a fair assessment of the contribution of physics integration beyond pure data-driven modeling. Specifically, three baseline ML models span a broad range of model complexity. The ridge polynomial regression model represents a low-complexity parametric model with limited nonlinear flexibility. The ANN provides a moderate level of complexity through multilayer nonlinear transformations but does not explicitly model sequential dependencies. The transformer represents a high-complexity architecture with attention-based sequence modeling.

### 2.3.2 Physics-based constitutive models

In addition, a physics-based constitutive model which purely relies on predefined physical laws and material parameters is included to assess the necessity of incorporating ML algorithm. In each fold of the polymer datasets, five material parameters are fixed and shared across all test samples. These fixed parameters are computed as the mean value of the corresponding values fitted by the training samples based on the nonlinear least squares. Specifically, $\bar{E}$ is used in the Hooke's law for the elastic region, $\bar{\varepsilon}_y$, $\bar{\sigma}_y$, $\bar{a}$, and $\bar{b}$ are used in the Voce hardening law. Therefore, the physics-based constitutive model for the polymer samples is defined as:

$$\sigma_p^t = \bar{E} * \varepsilon^t, \varepsilon < \bar{\varepsilon}_y$$
$$\sigma_p^t = \bar{\sigma}_y + \bar{a}(1 - e^{-\bar{b}(\varepsilon^t - \bar{\varepsilon}_y)}), \varepsilon \geq \bar{\varepsilon}_y$$
(11)

Similarly for each fold of the metal datasets, five material parameters are fixed and shared across all test samples. These fixed parameters are computed as the mean value of the corresponding values fitted by the training samples based on the nonlinear least squares. Specifically, $\bar{E}$ is used in the Hooke's law for the elastic region, $\bar{\varepsilon}_y$, $\bar{\sigma}_y$, $\bar{K}$, and $\bar{n}$ are used in the Hollomon's law. Therefore, the physics-based constitutive model for the metal samples is defined as:

$$\sigma_p^t = \bar{E} * \varepsilon^t, \varepsilon < \bar{\varepsilon}_y$$
$$\sigma_p^t = \bar{\sigma}_y + \bar{K}(\varepsilon - \bar{\varepsilon}_y)^{\bar{n}}, \varepsilon \geq \bar{\varepsilon}_y$$
(12)

Several error metrics, including MAPE and coefficient of determination ($R^2$), are used to evaluate the performance of the models. These error metrics are defined as follows:

$$\text{MAPE (\%)} = \frac{\sum_{i=1}^{n} \frac{|y_i - \hat{y}_i|}{y_i}}{n} \times 100 \tag{13}$$

$$R^2 = 1 - \frac{\sum_{i=1}^{n}(y_i - \hat{y}_i)^2}{\sum_{i=1}^{n}(y_i - \bar{y}_i)^2} \tag{14}$$

where n is the number of values, $y_i$ is the actual value, $\hat{y}_i$ is the predicted value, $\bar{y}_i$ is the mean of actual value.

To quantify the uncertainty of the reported mean values, we further compute the 95% confidence interval (CI) for each metric. The 95% CI is calculated as:

$$CI = \bar{x} \pm t^*\left(\frac{s}{\sqrt{n}}\right) \tag{15}$$

where $\bar{x}$ and $s$ is the mean value and standard deviation of a metric, $t^*$ is the critical value from the t-table that depends on the degrees if freedom.

## 2.4 Experimental design and data description

The Nylon and CF-ABS samples were fabricated via an FFF-based Ultimaker S3 printer (Ultimaker, Netherlands) with a 250 μm nozzle diameter. The diameters of the Nylon (Ultimaker, Netherlands) and CF-ABS (Push Plastic, USA) filaments are 2.85 mm diameter. All polymer samples were fabricated in accordance with ASTM D638 type V [46]. Uniaxial tensile tests were conducted on a universal test frame (AGS-X Series, Shimadzu, Japan) to characterize the stress-strain curves of the samples. The load capacity of the test frame was 10 kN. The test speed and sampling frequency were 2 mm/min and 10 Hz, respectively. A full factorial design of experiments (DOE) was developed to collect experimental data by varying critical process parameters. Table 2 and Table A.1 show the DOE for fabricating the Nylon and CF-ABS samples. Each material involved two process parameters, and each parameter was set at five numerical levels, resulting in a total of 25 samples per polymer dataset.

Table 2. Experimental design for the polymer and metal samples.

|  | Material | Parameter | Number of samples | Standard |
|---|---|---|---|---|
| Polymer | Nylon | Print speed (mm/s) Print temperature (°C) | 25 | ASTM D638 |
|  | CF-ABS | Print speed (mm/s) Print temperature (°C) | 25 |  |
| Metal | AlSi10Mg | Laser power (W) Scanning speed (mm/s) | 28 | ASTM E8/E8M |
|  | Ti6Al4V | Laser power (W) Scanning speed (mm/s) | 42 |  |

The AlSi10Mg dataset was published by the Pennsylvania State University [47]. All the samples were fabricated with a 3D Systems ProX DMP 320 L-PBF system machine (3D System, Inc, United States) using gas atomized LaserForm AlSi10Mg (A) metallic powder (3D System, Inc, United States) in accordance with ASTM E8/E8M [48]. Table 2 and Table B.1 show the DOE for fabricating the AlSi10Mg samples. In total, 28 samples were fabricated with varying laser powers and scanning speeds. The Ti6Ai4V dataset was also published by the Pennsylvania State University [49], all samples were fabricated accordance with the same ASTM standard and L-PBF system using LaserForm Ti Gr23(A) Ti-6Al-4V powder (3D Systems, Inc, United States). Table 2 and Table B.1 show the DOE of Ti6Al4V samples. In total, 42 samples were fabricated with varying laser powers and scanning speeds. After fabrication, uniaxial tensile tests were performed for the metal samples by an electromechanical load frame (Criterion Model 45, MTS Systems, Inc., United States) with maximum load of 10 kN and a strain rate of $3 \times 10^{-4}$/s.

Table 3 provides the divisions of the training and test datasets for 5-fold cross validation. The use of 5-fold cross-validation balances computational efficiency and statistical reliability. By cyclically rotating the training and test datasets, cross validation minimizes the potential bias that could arise from a single train/test split and provides a more comprehensive evaluation of model stability and performance across different subsets of data. Specifically, for each material, 20% of the total samples are selected as the training dataset, while the remaining 80% are the test dataset in each fold. Therefore, every sample is used once for training and four times for testing across the five folds, providing a robust assessment of model generalization.

Table 3. Divisions of the training and test datasets for each fold in 5-fold cross validation.

|  | Material | Training dataset | Test dataset |
|---|---|---|---|
| Polymer | Nylon | 20% of the dataset: 5 samples | Remaining 80% of the dataset: 20 samples |
|  | CF-ABS | 20% of the dataset: 5 samples | Remaining 80% of the dataset: 20 samples |
| Metal | AlSi10Mg | 20% of the dataset: 5 samples | Remaining 80% of the dataset: 23 samples |
|  | Ti6Al4V | 20% of the dataset: 8 samples | Remaining 80% of the dataset: 34 samples |

## 3. Results
### 3.1 Mechanical behaviors of additively manufactured polymers and metals

Fig. 4 shows the stress-strain curves of samples fabricated by two polymers (i.e., Nylon and CF-ABS) and two metals (i.e., AlSi10Mg and Ti6Al4V) with varying process parameters. To better distinguish mechanical behaviors of each material, the corresponding distributions of Young's modulus and ultimate tensile strength (UTS) are also included in Fig.4, highlighting differences in magnitude and variability within the same material.

In general, these four materials exhibit distinct mechanical behaviors, including yield strain, yield stress, stiffness, strength, and ductility. For the polymers, the CF-ABS samples exhibit higher stiffness and strength but lower ductility than the Nylon samples, primarily due to the reinforcing effect of carbon fibers, which restricts chain mobility and enhances load transfer. In contrast, the Nylon samples demonstrate oscillatory softening after yielding and longest plastic region among four materials. For the metals, the Ti6Al4V samples show significantly higher yield stress and young's modulus compared with the AlSi10Mg samples, consistent with its $\alpha - \beta$ titanium alloy microstructure that offers superior load-bearing capability, whereas the AlSi10Mg samples exhibit lower strength ductility due to its relatively brittle $Al - Si$ phase. When comparing the metal samples with the polymer samples, the metal samples exhibit stress levels in an order of magnitude higher and much steeper initial slopes than the polymer samples, indicating far greater stiffness and strength.

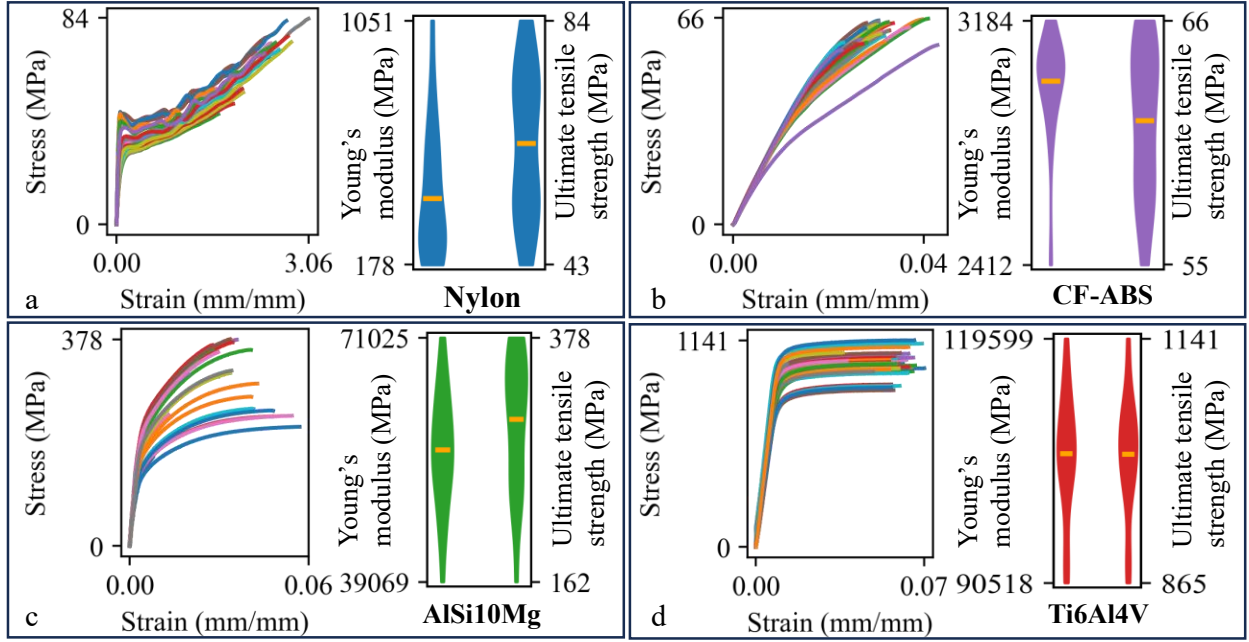

Fig. 4. Stress-strain curves, the distributions of Young's modulus and UTS of (a) Nylon, (b) CF-ABS, (c) AlSi10g, and (d) Ti6Al4V samples.

## 3.2 Predictive performance for additively manufactured polymer samples

Table 4. MAPEs between predicted and actual stress-strain curves of the polymer samples.

| | Nylon | | |
|---|---|---|---|
| | Elastic region | Plastic region | Whole curve |
| Non-segmental predictive model | - | - | 13.42% |
| Segmental predictive model without PIML architecture | 19.75% | 12.01% | 12.42% |
| Segmental predictive model with loss-based PIML architecture | **18.27%** Average $\lambda_{phys}$ = 0.9284 | 5.86% Average $\lambda_{phys}$ = 0.9978 | 6.53% |
| Segmental predictive model with activation-based PIML architecture | 18.52% | **4.87%** | **5.52%** |
| | CF-ABS | | |
| | Elastic region | Plastic region | Whole curve |
| Non-segmental predictive model | - | - | 23.90% |
| Segmental predictive model without PIML architecture | 23.41% | 12.14% | 15.36% |
| Segmental predictive model with loss-based PIML architecture | 13.69% Average $\lambda_{phys}$ = 0.9881 | **7.57%** Average $\lambda_{phys}$ = 0.9912 | 9.28% |
| Segmental predictive model with activation-based PIML architecture | **7.33%** | 8.59% | **8.25%** |

Table 4 shows a comparison of the MAPEs of different models between predicted and actual stress-strain curves of additively manufactured polymer samples (i.e., the Nylon and CF-ABS samples). To highlight the advantage of integrating segmentation and physics, two PIML architecture are compared with two

LSTM-based ML models, including the non-segmental predictive model and the segmental predictive model without PIML. The non-segmental predictive model treated the stress-strain curve as a single, continuous curve without any segmentation. The segmental predictive model without PIML adopted the same segmentation strategy, but without embedding any physical law. Except for the MAPEs, average $\lambda_{phys}$ across all folds is included for the segmental predictive model with loss-based PIML. Since each average $\lambda_{phys}$ is close to 1, the auxiliary loss $L_{\text{auxiliary}}$ successfully completes its role. For both of the polymer samples, the segmental predictive model with PIML outperforms the non-segmental predictive model and the segmental predictive model without PIML. Overall, the activation-based PIML model achieves the best predictive performance with an average MAPE of 5.52% for the Nylon samples and 8.25% for the CF-ABS samples, respectively. However, an opposite trend is observed in the elastic region of the Nylon samples and the plastic region of the CF-ABS samples, where the loss-based PIML architecture slightly outperforms the activation-based PIML architecture. The reason is the elastic regions of the Nylon samples exhibit a less distinct linear elastic behavior, enforcing the Hooke's law through a hard constraint (i.e., activation function) limits the flexibility of the model. In contrast, with a soft constraint (i.e., loss function), the model can better accommodate the nonlinearity in the elastic regions of the Nylon samples.

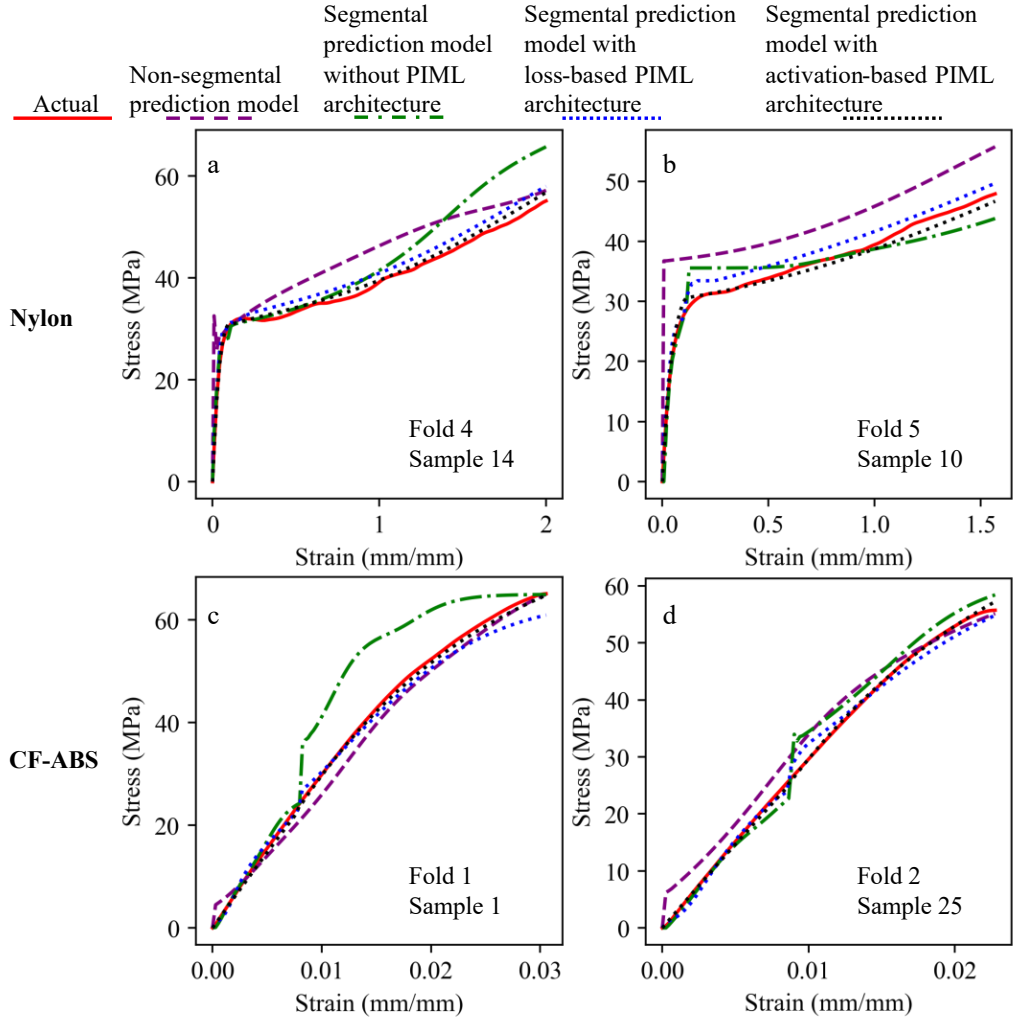

Fig. 5. Actual versus predicted stress-strain curves of additively manufactured polymer samples using different models. The segmental predictive model with activation-based PIML architecture outperforms the other models. Nylon: (a) sample 14 in Fold 4 and (b) sample 10 in Fold 5. CF-ABS: (a) sample 1 in Fold 1 and (b) sample 25 in Fold 2.

Fig. 5 shows the actual versus predicted stress-strain curves of additively manufactured polymer samples by different models (i.e., Figs. 5a and 5b for the Nylon samples, Figs. 5c and 5d for the CF-ABS samples). Overall, compared with the actual curves (red lines), the predictions from the non-segmental predictive models (purple lines) and the segmental predictive models without PIML architecture (green lines) show noticeable deviations in both elastic and plastic regions of the polymer samples. In contrast, the predicted curves by two PIML architectures closely follow the actual ones. The activation-based PIML architectures (black lines) achieve the best predictive performance and exhibit the smoothest transitions between elastic and plastic regions. Specifically, for Nylon sample 14 in Fold 4 (Fig. 5a), the predicted curve by the non-segmental predictive model shows an unrealistic abrupt slope change near yield point and large deviations in the plastic region. Although all other models perform reasonably well in the elastic region, only the segmental predictive model with activation-based PIML architecture yields a curve that remains highly in agreement with the actual curve in the plastic region. Similarly, for Nylon sample 10 in Fold 5 (Fig. 5b), the predicted curve by the segmental predictive model with activation-based PIML architecture consistently follow the actual curve, especially for the yield point; the other models all overestimate the yield stress. For CF-ABS sample 1 in Fold 1 (Fig. 5c), the non-segmental predictive model performs poorly at the beginning of the elastic region, while the segmental model without PIML architecture shows noticeable deviation in the plastic region. The segmental predictive model with loss-based PIML architecture underestimates the UTS, whereas the segmental predictive model with activation-based PIML architecture provides the most accurate prediction across the entire stress-strain curve. Similar trends can also be observed in CF-ABS sample 25 in Fold 2 (Fig. 5d).

Table 5. MAPEs between predicted and actual Young's modulus and UTS of polymer samples.

| | Nylon | |
|---|---|---|
| | Young's modulus | UTS |
| Non-segmental predictive model | 38.98% | 7.10% |
| Segmental predictive model without PIML architecture | 34.68% | 17.79% |
| Segmental predictive model with loss-based PIML architecture | 23.48% | 5.69% |
| Segmental predictive model with activation-based PIML architecture | **20.54%** | **3.81%** |
| | CF-ABS | |
| | Young's modulus | UTS |
| Non-segmental predictive model | 18.54% | 8.32% |
| Segmental predictive model without PIML architecture | 9.33% | 6.90% |
| Segmental predictive model with loss-based PIML architecture | 5.23% | **5.19%** |
| Segmental predictive model with activation-based PIML architecture | **4.41%** | 6.02% |

In addition, the MAPEs of two key mechanical properties, which are Young's modulus and UTS, are also reported to quantitatively demonstrate the improved physical consistency of two PIML architectures. As shown in Table 5, for the Nylon samples, the segmental predictive model with activation-based PIML architecture achieves the lowest MAPE of Young's modulus and UTS of 20.54% and 3.81%, respectively. For the CF-ABS samples, the segmental predictive model with activation-based PIML architecture achieves the lowest MAPE of 4.41% for predicting Young's modulus, while the segmental predictive model with loss-based PIML architecture achieves the lowest MAPE of 5.19% for predicting UTS.

Fig. 6 illustrates the training evolution of different loss components for the loss-based and activation-based PIML architectures within the plastic region of the Nylon dataset in Fold 1. It can be seen that as training progresses, all the losses reach relatively low levels at the 10th training epoch, indicating stable convergence of two PIML architectures. The auxiliary loss (Fig. 6c) is designed to constrain the learnable physics weighting coefficient $\lambda_{physics}$ to remain close to 1 (Fig. 6d), ensuring that the contributions of the data-driven and physics-informed terms are balanced during training. Therefore, after 5 training epochs, the auxiliary loss is almost equal to 0, ensuring that it no longer interferes with the training process.

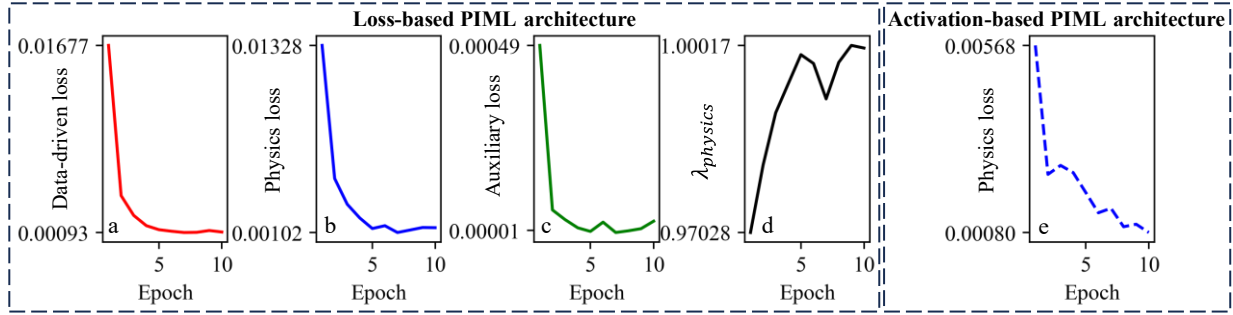

Fig. 6. Training evolution of (a) data-driven, (b) physics, and (c) auxiliary losses as well as (d) the learnable physics weighting coefficient $\lambda_{physics}$ for the loss-based PIML architecture within the plastic region of the Nylon dataset in Fold 1. Training evolution of (e) physics loss for the activation-based PIML architecture within the plastic region of the Nylon dataset in Fold 1.

### 3.3 Predictive performance for additively manufactured metal samples

Table 6. MAPEs between predicted and actual stress-strain curves of metal samples.

| | AlSi10Mg | | |
|---|---|---|---|
| | Elastic region | Plastic region | Whole curve |
| Non-segmental predictive model | - | - | 26.09% |
| Segmental predictive model without PIML | 28.99% | 23.10% | 25.76% |
| Segmental predictive model with loss-based PIML | 20.90% Average $\lambda_{phys}$ = 0.9912 | 17.04% Average $\lambda_{phys}$ = 0.9987 | 17.03% |
| Segmental predictive model with activation-based PIML | **18.55%** | **16.84%** | **16.51%** |
| | Ti6Al4V | | |
| | Elastic region | Plastic region | Whole curve |
| Non-segmental predictive model | - | - | 17.93% |
| Segmental predictive model without PIML | 23.07% | 6.12% | 15.28% |
| Segmental predictive model with loss-based PIML | 16.39% Average $\lambda_{phys}$ = 0.9996 | 5.61% Average $\lambda_{phys}$ = 0.9939 | 11.63% |
| Segmental predictive model with activation-based PIML | **14.82%** | **5.19%** | **10.58%** |

Table 6 shows a comparison of the MAPEs of different models between predicted and actual stress-strain curves of additively manufactured metal samples (i.e., the AlSi10Mg and Ti6Al4V samples). Except for the MAPEs, average $\lambda_{phys}$ across all folds are included for the segmental predictive model with loss-based PIML architecture. Since each average $\lambda_{phys}$ is close to 1, the auxiliary loss $L_{\text{auxiliary}}$ successfully completes its role. Across both metal samples, the segmental predictive model with activation-based PIML architecture consistently achieves the best predictive performance in the elastic region, plastic region, and whole stress-strain curve. This trend is expected since metallic materials generally follow well-defined physical laws because of the consistent dislocation motion. Therefore, the hard constraint embedded in the activation function can help the model learn the physics more effectively. Specifically, for the AlSi10Mg samples, the segmental predictive model with activation-based PIML architecture achieves the best predictive performance with an average MAPE of 18.55%, 16.84%, and 16.51% for the elastic region, plastic region, and whole stress-strain curve, respectively, followed by the segmental predictive model with loss-based PIML architecture (i.e., 20.90%, 17.04%, 17.03%), the segmental predictive model without PIML architecture (i.e., 28.99%, 23.10%, 25.76%), and the worst non-segmental predictive model with the highest MAPE of 26.09% for the whole stress-strain curve. Similarly, for the Ti6Al4V samples, the segmental predictive model with activation-based PIML architecture yields the lowest MAPE of 14.82% for the elastic region, 5.19% for the plastic region, and 10.58% for the whole stress-strain curve.

Fig. 7 shows the actual versus predicted stress-strain curves of additively manufactured metal samples (i.e., Figs. 7a and 7b for the AlSi10Mg samples, Figs. 7c and 7d for the Ti6Al4V samples) by different models. Overall, compared with the actual curves (red lines), the predictions from the non-segmental predictive model (purple lines) and the segmental predictive model without PIML architecture (green lines) show noticeable deviations in both elastic and plastic regions. In contrast, the predicted curves by two PIML architectures closely follow the actual ones. The segmental predictive model with activation-based PIML architecture (black lines) achieves the best predictive performance and exhibits the smoothest transitions between elastic and plastic regions. Specifically, for the AlSi10Mg sample 16 in Fold 3 (Fig. 7a), the non-segmental predictive model shows the worst predictive performance in the plastic region among all models, while the segmental predictive model without PIML architecture exhibits the worst predictive performance in the elastic region among all models. The other two PIML architectures both perform well in the elastic region, while the predictions by the activation-based PIML architecture shows better agreement with the actual stress-strain curve in the plastic region. For the AlSi10Mg sample 14 in Fold 4 (Fig. 7b), in the plastic region, both the predicted stress-strain curves by the non-segmental predictive model and segmental predictive model with loss-based PIML architecture exhibit unrealistic stress drops near the end of the curve, indicating unstable predictions without physical consistency. In contrast, the segmental predictive model with activation-based PIML architecture provides the most accurate prediction across the whole stress-strain curve. For the Ti6Al4V sample 19 in Fold 2 (Fig. 7c), the differences in predictive performance among different models are primarily observed in the plastic region. The segmental predictive model without PIML architecture shows the largest deviation, followed by the non-segmental predictive model. The segmental predictive model with activation-based PIML architecture achieves the best predictive performance in the plastic region. For the Ti6Al4V sample 29 in Fold 5 (Fig. 7d), more obvious differences appear across models in both the elastic and plastic regions. The non-segmental predictive model significantly underestimates stress throughout the stress-strain curve and predicts an unrealistic plateau at the beginning of the elastic region. The segmental predictive model with loss-based PIML architecture improves predictive performance but still underestimates the stresses in the plastic region. In contrast, the predicted curve by the segmental predictive model with activation-based PIML architecture closely follows the actual stress-strain curve, accurately capturing both the yield point and the subsequent hardening behavior in the elastic region.

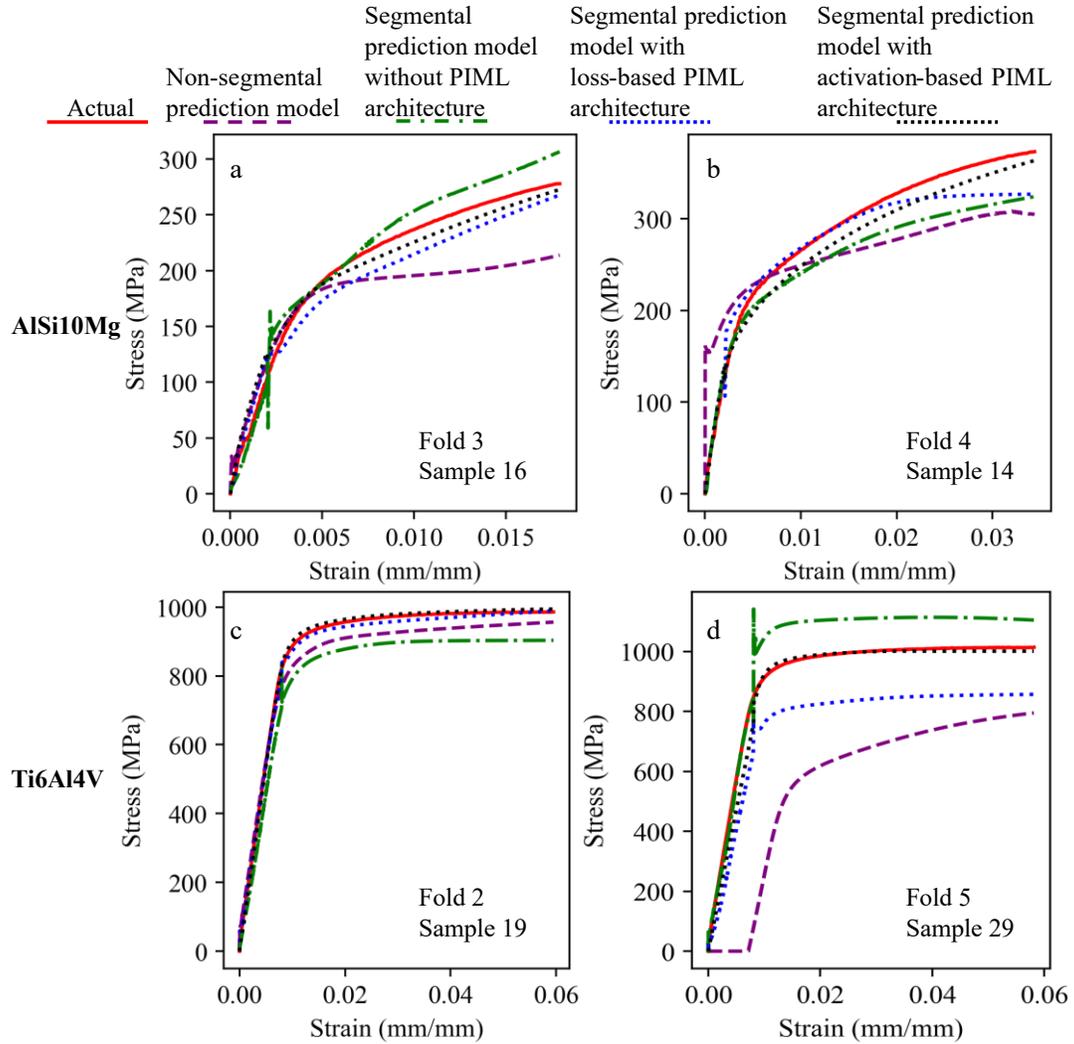

Fig. 7. Actual versus predicted stress-strain curves of additively manufactured metal samples using different models. The segmental predictive model with activation-based PIML architecture outperforms the other models. AlSi10Mg: (a) sample 16 in Fold 3 and (b) sample 14 in Fold 4. Ti6Al4V: (a) sample 19 in Fold 2 and (b) sample 29 in Fold 5.

In addition, the MAPEs of two key mechanical properties, which are Young's modulus and UTS, are also reported to quantitatively demonstrate the improved physical consistency of two PIML architectures. As shown in Table 7, for the AlSi10Mg samples, the segmental predictive model with loss-based PIML architecture both achieves the lowest MAPE of Young's modulus and UTS of 11.24% and 16.89%, respectively. For the Ti6Al4V samples, the segmental predictive model with activation-based PIML architecture both achieves the lowest MAPE of Young's modulus and UTS of 5.65% and 4.98%, respectively.

Table 7. MAPEs between predicted and actual Young's modulus and UTS of the metal samples.

| | AlSi10Mg | |
|---|---|---|
| | Young's modulus | Ultimate strength |
| Non-segmental predictive model | 21.64% | 19.59% |
| Segmental predictive model without PIML | 25.47% | 24.16% |
| Segmental predictive model with loss-based PIML | **11.24%** | **16.89%** |
| Segmental predictive model with activation-based PIML | 16.44% | 18.53% |
| | Ti6Al4V | |
| | Young's modulus | Ultimate strength |
| Non-segmental predictive model | 12.27% | 8.77% |
| Segmental predictive model without PIML | 10.78% | 5.59% |
| Segmental predictive model with loss-based PIML | 6.81% | 5.39% |
| Segmental predictive model with activation-based PIML | **5.65%** | **4.98%** |

Fig. 8 illustrates the training evolution of different loss components for the loss-based and activation-based PIML architectures within the elastic region of the AlSi10Mg dataset in Fold 1. It can be seen that as training progresses, all the losses reach relatively low levels at the 10th training epoch, indicating stable convergence of the two PIML architectures. The auxiliary loss (Fig. 8c) is designed to constrain the learnable physics weighting coefficient $\lambda_{physics}$ to remain close to 1 (Fig. 8d), ensuring that the contributions of the data-driven and physics-informed terms are balanced during training. Therefore, after 4 training epochs, the auxiliary loss is almost equal to 0, ensuring that it no longer interferes with the training process.

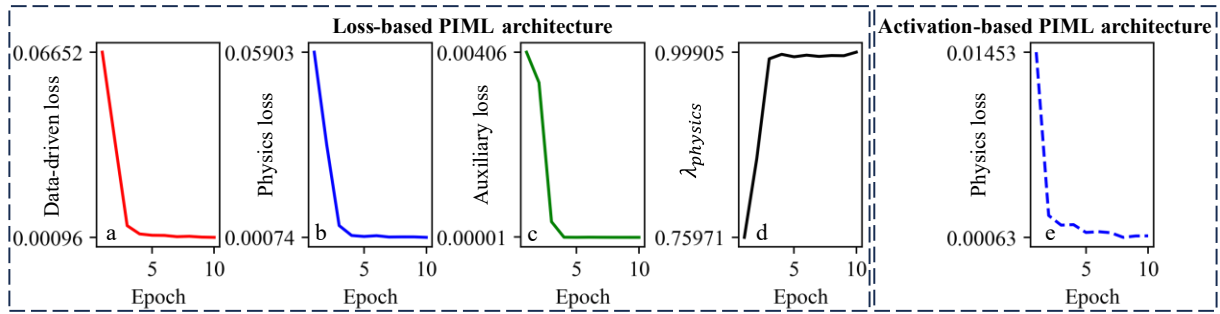

Fig. 8. Training evolution of (a) data-driven, (b) physics, and (c) auxiliary losses as well as (d) the learnable physics weight $\lambda_{physics}$ for the loss-based PIML architecture within the elastic region of AlSi10Mg dataset in Fold 1. Training evolution of (e) physics loss for the activation-based PIML architecture within the elastic region of AlSi10Mg dataset in Fold 1.

### 3.4 Average predictive performance comparison

Table 8 summarizes the average predictive performance of different models across four datasets (i.e., Nylon, CF-ABS, AlSi10Mg, and Ti6Al4V datasets), with 95% CI reported to quantify statistical uncertainty. Compared with three additional ML models and a physics-based constitutive model, both PIML architectures demonstrate substantially improved predictive performance.

Among the three additional ML models, the ridge polynomial regression model exhibits the weakest predictive performance with MAPE of 40.56±2.96% and $R^2$ of 0.33±0.20, indicating this simple regression model has limited capacity to capture the nonlinear stress-strain behaviors. Although ANN and transformer improve predictive performance, their errors remain significantly higher than those of the PIML architectures. Although the physics-based constitutive model achieves relatively competitive MAPE of 18.33±0.82%, the $R^2$ of 0.34±0.06 indicates that the physics-based constitutive model can only capture the general trend of the stress-strain curve. The limited flexibility of the constitutive mode restricts its ability to model complex deviations from idealized deformation behavior. In contrast, the segmental predictive models with loss-based or activation-based architectures effectively balance physical constraints and data adaptability. The segmental predictive model with loss-based PIML architecture achieves an average MAPE of 11.38±0.90% and an average $R^2$ of 0.80±0.06), while the activation-based PIML architecture further improves predictive performance to the lowest MAPE of 10.46±0.81% with the highest $R^2$ of 0.82±0.05. The low 95% CIs indicate stable predictive performance across folds and materials.

Table 8. Average predictive performance of PIML, ML, and physics-based constitutive models.

|  | MAPE±95% CI | $R^2$±95% CI |
|---|---|---|
| Physics-based constitutive model | 18.33±0.82% | 0.34±0.06 |
| Ridge polynomial regression model | 40.56±2.96% | 0.33±0.20 |
| ANN | 18.86±1.48% | 0.73±0.05 |
| Transformer | 22.33±1.79% | 0.70±0.07 |
| Segmental predictive model with loss-based PIML architecture | 11.38±0.90% | 0.80±0.06 |
| Segmental predictive model with activation-based PIML architecture | **10.46±0.81%** | **0.82±0.05** |

In general, both loss-based and activation-based PIML architectures effectively integrate physical laws into the ML model, enhancing the predictive performance and physical consistency of stress-strain behaviors compared with ML model. However, their mechanisms and advantages differ. The loss-based PIML architecture introduces physical laws as a soft constraint through an additional loss term, offering greater flexibility and stability when data noise or slight nonlinearity exists. In contrast, the activation-based PIML architecture embeds the physical laws directly into the network structure, enforcing a hard constraint that ensures strict physical consistency. This design generally yields higher predictive performance and smoother predictions, particularly for data that are closely following well-defined physical laws. Therefore, the activation-based PIML architecture demonstrates stronger generalization and physical fidelity, while the loss-based PIML architecture remains advantageous for data with complex nonlinear behavior.

## 4. Conclusions

In this study, a material-specific and region-specific PIML framework was developed to predict the stress-strain curves of additively manufactured polymers and metals. The framework was demonstrated on two additively manufactured polymer datasets (i.e., Nylon and CF-ABS) and two additively manufactured metal datasets (i.e., AlSi10Mg and Ti6Al4V). A ridge polynomial regression model was first used to predict the yield point based on the process parameters so that each stress-strain curve was segmented into an elastic region and a plastic region. For the elastic region, Hooke's law was leveraged by the PIML framework. For the plastic region, the Voce hardening law and Hollomon's law were leveraged by the PIML framework for the polymer and metal samples, respectively. Specifically, these physical laws were embedded in either an additional physics loss function or an activation function of the output layer. Two PIML architectures were further compared with two LSTM-based ML models (i.e., the non-segmental predictive model and the segmental predictive model without PIML architecture), three additional ML models (i.e., ridge polynomial regression model, ANN, and transformer) and a physics-based constitutive model. Experimental results showed that two PIML architectures consistently outperformed all the other models on two additively manufactured polymer datasets and two additively manufactured metal datasets. Furthermore, the segmental predictive model with activation-based PIML architecture achieves the best predictive performance with the lowest MAPE of 10.46±0.81% and the highest $R^2$ of 0.82±0.05 across four datasets. In future work, additional physical laws designed for other critical regions will be incorporated into the PIML framework.

**Acknowledgments**
This research is sponsored by the Defense Advanced Research Projects Agency. The content of the information does not necessarily reflect the position or the policy of the Government. No official endorsement should be inferred. We would like to thank Dr. Qingyang Liu for collecting the raw data for the polymer datasets.

**Data Availability Statement**
The dataset generated and supporting the findings of this article are available upon reasonable request from the corresponding author.

**Appendix A. DOE of polymer samples**

Table A.1 DOE for fabricating the Nylon and CF-ABS samples via FFF.

| Nylon | | | CF-ABS | | |
|---|---|---|---|---|---|
| Sample ID | Print temperature (°C) | Print speed (mm/s) | Sample ID | Print temperature (°C) | Print speed (mm/s) |
| 1 | 220 | 10 | 1 | 200 | 10 |
| 2 | 220 | 20 | 2 | 200 | 30 |
| 3 | 220 | 30 | 3 | 200 | 50 |
| 4 | 220 | 40 | 4 | 200 | 70 |
| 5 | 220 | 50 | 5 | 200 | 90 |
| 6 | 230 | 10 | 6 | 220 | 10 |
| 7 | 230 | 20 | 7 | 220 | 30 |
| 8 | 230 | 30 | 8 | 220 | 50 |
| 9 | 230 | 40 | 9 | 220 | 70 |
| 10 | 230 | 50 | 10 | 220 | 90 |
| 11 | 240 | 10 | 11 | 240 | 10 |
| 12 | 240 | 20 | 12 | 240 | 30 |
| 13 | 240 | 30 | 13 | 240 | 50 |
| 14 | 240 | 40 | 14 | 240 | 70 |
| 15 | 240 | 50 | 15 | 240 | 90 |
| 16 | 250 | 10 | 16 | 260 | 10 |
| 17 | 250 | 20 | 17 | 260 | 30 |
| 18 | 250 | 30 | 18 | 260 | 50 |
| 19 | 250 | 40 | 19 | 260 | 70 |
| 20 | 250 | 50 | 20 | 260 | 90 |
| 21 | 260 | 10 | 21 | 280 | 10 |
| 22 | 260 | 20 | 22 | 280 | 30 |
| 23 | 260 | 30 | 23 | 280 | 50 |
| 24 | 260 | 40 | 24 | 280 | 70 |
| 25 | 260 | 50 | 25 | 280 | 90 |

# Appendix B. DOE of metal samples

Table B.1 DOE for fabricating the AlSi10Mg and Ti6Al4V samples via L-PBF.

| AlSi10Mg | | | Ti6Al4V | | |
| --- | --- | --- | --- | --- | --- |
| Sample ID | Laser power (W) | Scanning speed (mm/s) | Sample ID | Laser power (W) | Scanning speed (mm/s) |
| 1 | 60 | 250 | 1 | 275 | 800 |
| 2 | 160 | 250 | 2 | 275 | 760 |
| 3 | 160 | 800 | 3 | 275 | 720 |
| 4 | 260 | 250 | 4 | 275 | 680 |
| 5 | 260 | 800 | 5 | 275 | 640 |
| 6 | 260 | 1350 | 6 | 275 | 600 |
| 7 | 360 | 250 | 7 | 275 | 840 |
| 8 | 360 | 800 | 8 | 275 | 880 |
| 9 | 360 | 1350 | 9 | 275 | 920 |
| 10 | 360 | 1900 | 10 | 275 | 960 |
| 11 | 460 | 250 | 11 | 275 | 1000 |
| 12 | 460 | 800 | 12 | 175 | 800 |
| 13 | 460 | 1350 | 13 | 195 | 800 |
| 14 | 460 | 1900 | 14 | 215 | 800 |
| 15 | 460 | 2450 | 15 | 235 | 800 |
| 16 | 110 | 250 | 16 | 255 | 800 |
| 17 | 110 | 500 | 17 | 295 | 800 |
| 18 | 160 | 500 | 18 | 315 | 800 |
| 19 | 260 | 500 | 19 | 335 | 800 |
| 20 | 360 | 500 | 20 | 355 | 800 |
| 21 | 460 | 500 | 21 | 375 | 800 |
| 22 | 210 | 250 | 22 | 135 | 400 |
| 23 | 210 | 500 | 23 | 205 | 600 |
| 24 | 210 | 800 | 24 | 345 | 1000 |
| 25 | 210 | 1100 | 25 | 415 | 1200 |
| 26 | 260 | 1100 | 26 | 485 | 1400 |
| 27 | 360 | 1100 | 27 | 500 | 1480 |
| 28 | 460 | 1100 | 28 | 155 | 400 |
| | | | 29 | 235 | 600 |
| | | | 30 | 395 | 1000 |
| | | | 31 | 475 | 1200 |
| | | | 32 | 500 | 1290 |
| | | | 33 | 115 | 400 |
| | | | 34 | 175 | 600 |
| | | | 35 | 295 | 1000 |
| | | | 36 | 355 | 1200 |
| | | | 37 | 475 | 1600 |
| | | | 38 | 75 | 400 |
| | | | 39 | 115 | 600 |
| | | | 40 | 195 | 1000 |
| | | | 41 | 235 | 1200 |
| | | | 42 | 315 | 1600 |